\documentclass[10pt,twocolumn,letterpaper]{article}

\usepackage{cvpr}
\usepackage{times}
\usepackage{epsfig}
\usepackage{graphicx}
\usepackage{amsmath}
\usepackage{amssymb}

\usepackage[pagebackref]{hyperref}
\usepackage[flushleft]{threeparttable}
\usepackage{booktabs} 
\usepackage{caption}

\usepackage{multirow}
\usepackage{tabularx}
\usepackage{xcolor}
\usepackage{algorithm}
\usepackage{algpseudocode}

\newcolumntype{M}{>{$}c<{$}}
\newcolumntype{C}{>{\centering\arraybackslash}p}
\newcolumntype{Y}{>{\centering\arraybackslash}X}

\newcommand{\specialcell}[2][c]{%
  \begin{tabular}[#1]{@{}c@{}}#2\end{tabular}}

\newcommand{\leftcell}[2][l]{%
  \begin{tabular}[#1]{@{}l@{}}#2\end{tabular}}

\begin{document}

\title{ViT Unified: Joint Fingerprint Recognition and Presentation Attack Detection}

\author{Steven A. Grosz,  Kanishka P. Wijewardena, Anil K. Jain\\
Michigan State University\\
{\tt\small groszste@msu.edu, wijewar2@msu.edu, jain@msu.edu}}

\maketitle
\thispagestyle{empty}

\begin{abstract}
   A secure fingerprint recognition system must contain both a presentation attack (i.e., spoof) detection and recognition module in order to protect users against unwanted access by malicious users. Traditionally, these tasks would be carried out by two independent systems; however, recent studies have demonstrated the potential to have one unified system architecture in order to reduce the computational burdens on the system, while maintaining high accuracy. In this work, we leverage a vision transformer architecture for joint spoof detection and matching and report competitive results with state-of-the-art (SOTA) models for both a sequential system (two ViT models operating independently) and a unified architecture (a single ViT model for both tasks). ViT models are particularly well suited for this task as the ViT's global embedding encodes features useful for recognition, whereas the individual, local embeddings are useful for spoof detection. We demonstrate the capability of our unified model to achieve an average integrated matching (IM) accuracy of 98.87\% across LivDet 2013 and 2015 CrossMatch sensors. This is comparable to IM accuracy of 98.95\% of our sequential dual-ViT system, but with $\sim$50\% of the parameters and $\sim$58\% of the latency.
\end{abstract}

\begin{figure}[!t]
\includegraphics[width=\linewidth]{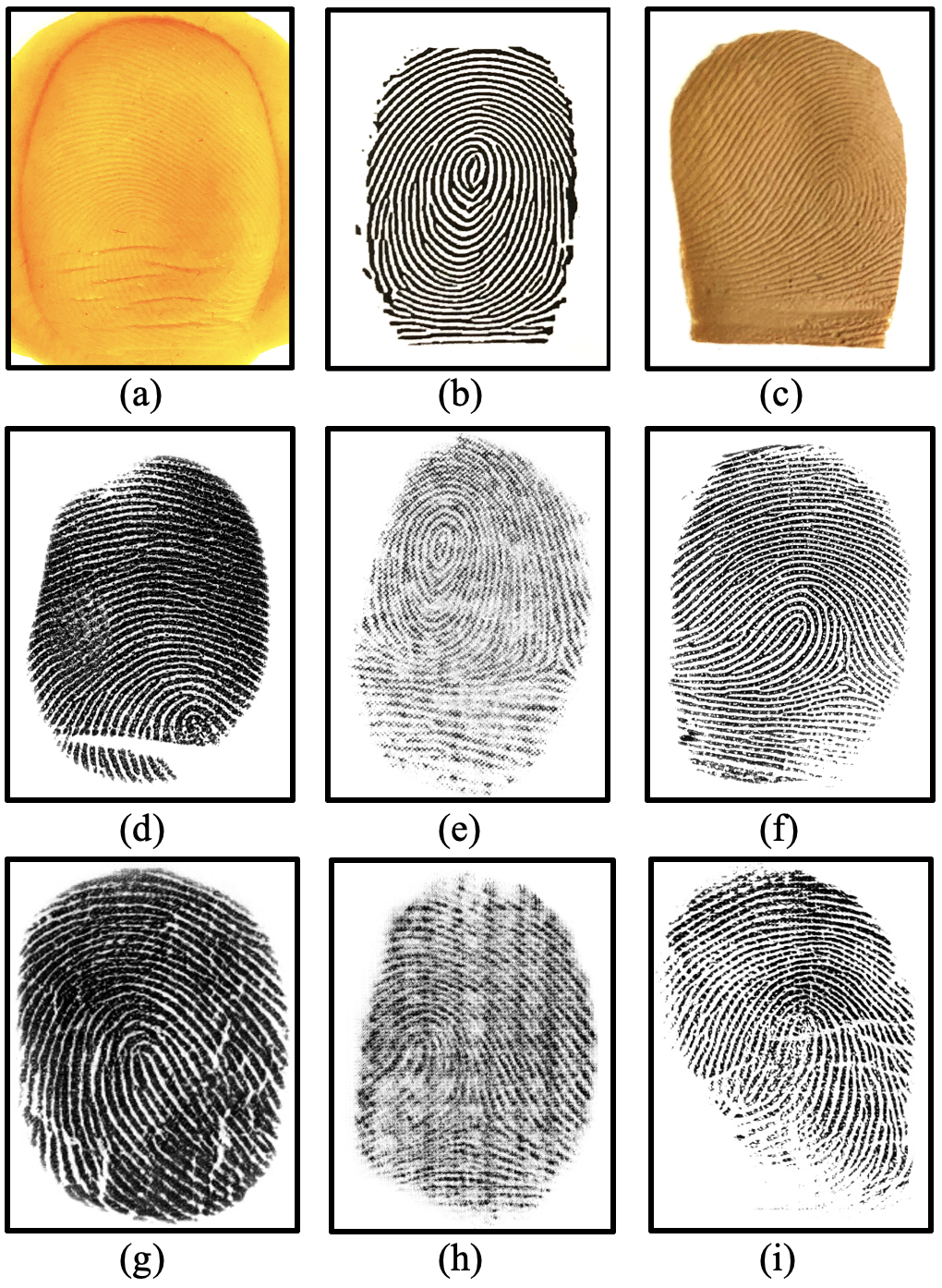} 
\caption{Example (a) Play-Doh, (b) printed paper, and (c) latex fabricated spoofs, their corresponding impressions (d-f) captured on a CrossMatch Guardian200 fingerprint reader, and (g-i) synthetic spoofs generated by SpoofGAN~\cite{grosz2022spoofgan}.}
\label{fig:ex_spoofs}
\end{figure}

\section{Introduction}
Secure biometric authentication systems are integral for proper identity verification in  various applications, such as access control, international travel and financial transactions \cite{jain_intro_biometrics}. Such systems not only require highly accurate identification, but also protection against intruders attempting to circumvent the normal operation of the biometric system; for example, through presentation attacks.
    
For the purposes of identity impersonation, the most commonly used presentation attack involves using a synthetic material to create an artifact  (commonly referred to as a ``spoof") of the fingerprint of an enrolled user. Figure~\ref{fig:ex_spoofs} shows some examples of such artifacts. Indeed, there have been reports of successful  presentation attacks on fingerprint access control systems, including smartphone unlock \cite{marasco_ross_antispoofing}. To combat these problems, software-based fingerprint spoof detection solutions have been tested and refined over the years and used in fingerprint recognition systems. Some of these software-based spoof detection systems have been evaluated in the Fingerprint Liveness Detection Competitions (LivDet), dating back to 2009 \cite{livdet_2009}. While the LivDet competitions prior to 2019 focused solely on fingerprint liveness detection, LivDet 2019 and LivDet 2021 have focused on both fingerprint liveness detection and identity verification systems operating in parallel and outputting a combined metric score~\cite{livdet_2019,livdet_2021}. The combined score is a proxy for evaluating the entire fingerprint recognition system in terms of its capability to detect both presentation attacks and zero-effort impostor attacks and only authenticate genuine, bona fide fingerprints.

Recent improvements in fingerprint matching systems have seen an increased use of deep network embeddings instead of and in addition to minutiae-based representations \cite{engelsma_deepprint,grosz_afrnet}. These fixed-length fingerprint embeddings are significantly faster compared to minutiae-based representations for both verification and large-scale search. These deep network embeddings are also more secure as they are less susceptible to inversion attacks that attempt to reconstruct the source fingerprint image from an enrolled template \cite{wijewardena_template}. For these reasons, we limit our exploration to deep network-based fingerprint representations in a joint fingerprint spoof detection and verification system.

More specifically, this paper presents the following contributions: 
\begin{enumerate}
    \item Competitive joint fingerprint presentation attack detection (PAD) and matching accuracy with significant reduction in latency compared to previous state-of-the-art (SOTA) methods.   
    \item Joint training of PAD and recognition with a unified Vision Transformer (ViT) architecture, achieving a reduction in $\sim$50\% of the parameters and $\sim$58\% of the latency compared to the use of individual ViT models for both tasks.
    \item Analysis of salient intermediate ViT features for fingerprint PAD and recognition.
    \item Extensive experimental results evaluating individual fingerprint PAD and recognition performance, as well as end-to-end joint performance.
    \item A comparison in the trade-off between accuracy and latency between sequentially operated (cascaded) joint spoof detection and verification systems and a unified system.
\end{enumerate}    
    
\begin{figure}[!t]
\includegraphics[width=\linewidth]{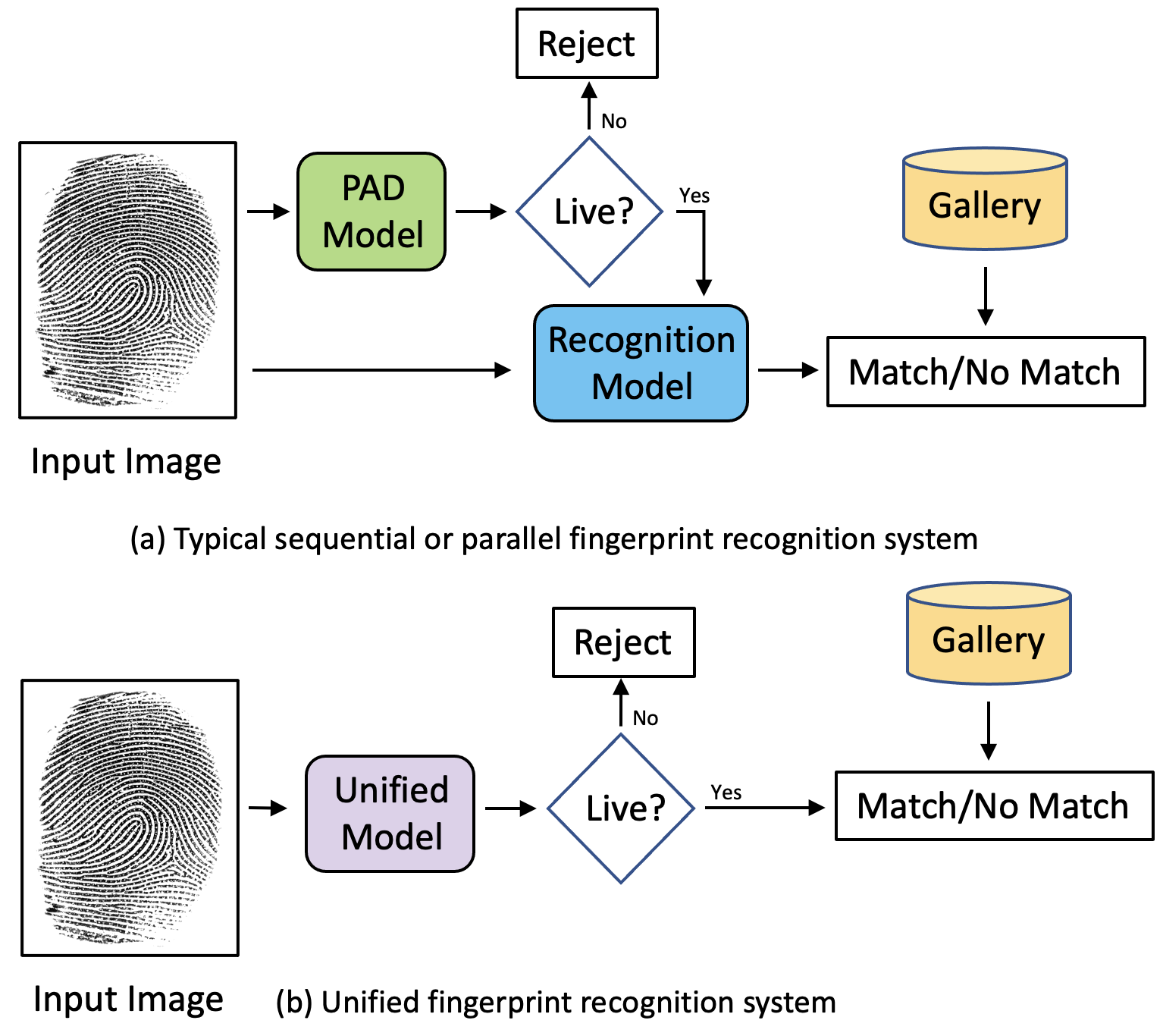} 
\caption{Comparison of (a) typical fingerprint recognition systems which employs separate PAD and recognition models vs. (b) a unified fingerprint recognition architecture where the two tasks are performed via a single model.}
\label{fig:overview}
\end{figure}

\begin{table*}
\caption{Summary of prior joint fingerprint presentation attack detection and matching studies.}
\begin{center}
{\footnotesize{
\begin{tabular}{l|l|l}
\noalign{\hrule height 1.5pt}
\textbf{Model} & \textbf{Approach} & \textbf{Results}\\
\noalign{\hrule height 1.0pt}
\leftcell{Fingerprint Liveness\\Detection Competions\\(LivDet 2019~\cite{livdet_2019} \& 2021~\cite{livdet_2021})} & \leftcell{Multiple submissions based on hand-crafted, hybrid\\ and/or deep learning.} & \leftcell{LivDet2021: IM Accuracy of 97.17\% and 99.11\%\\on GreenBit and Dermalog readers, respectively.}\\
\hline
\leftcell{Dual-Head MobileNet\\(DHM)~\cite{popli_unified}} & \leftcell{Utilized a multi-branch MobilenetV2 comprising of a\\spoof detection head and a fingerprint matching head.} & \leftcell{0.28\% ACE on CrossMatch data from LivDet2015.\\0.75\% FRR @ FAR 0.1\% on FVC 2002 DB1A.}\\
\hline
\leftcell{ONNS+SlimResCNN\\w/ score-level fusion~\cite{zhang_fusion}} & \leftcell{Using Octantal Nearest-Neighborhood Structure for\\fingerprint matching, and SlimResCNN for liveness\\detection and generating a score-level fusion vector.} & \leftcell{96.88\% overall accuracy on LivDet2019 across \\GreenBit, Orcanthus and Digital Persona readers.}\\
\noalign{\hrule height 1.5pt}
\end{tabular}
}}
\end{center}
\label{table_summary}
\end{table*} 

\section{Related Work}
Here we briefly discuss the prior literature in fingerprint liveness detection and joint fingerprint PAD and recognition.

\subsection{Fingerprint Liveness Detection}
Initial research into fingerprint recognition focused solely on improving the matching accuracy to reliably accept genuine fingerprint pairs (two fingerprint images belonging to the same finger) and rejecting imposter fingerprint pairs (two fingerprint images belonging to two different fingers). However, once fingerprint recognition accuracy began to saturate (now achieving an Equal Error Rate of 0.01\% on benchmark datasets \cite{fvc_ongoing}), discussions regarding the importance of anti-spoofing came to the forefront. Matsumoto et al. \cite{matsumoto_gummy} published their findings in 2002, where they showed that efforts to enroll spoof fingerprints in 11 types of fingerprint systems resulted in all systems accepting the spoof fingerprints in their verification procedures with a probability of more than 67\%. Several other works published during this time showed that spoofing attacks at the sensor-level could be carried out without prior knowledge of the template extraction or matching procedures \cite{galbally_survey_liveness}. In their survey on fingerprint anti-spoofing technologies, Marasco and Ross \cite{marasco_ross_antispoofing} divided anti-spoofing into hardware-based vs. software-based solutions, with the latter being the more affordable and often the preferred option. In this approach, the images captured by the sensor are subjected to further image processing, with a focus on either the dynamic behaviors of live fingertips such as ridge distortion or perspiration, or static characteristics such as the textural characteristics, ridge frequencies, or elastic properties of the skin. Here, implementing anti-spoofing mechanisms based on the static characteristics of an image is cheaper and faster. 

To further research into fingerprint anti-spoofing based on the static characteristics of an image, the Fingerprint Liveness Detection competition was proposed in 2009 \cite{livdet_2009}, dedicated to designing algorithms that could help differentiate between live fingerprints and spoof fingerprints. While there have been several spoof detection studies based on hand-crafted features~\cite{Ghianni_LPQ,Gragniello_Weber,gragniello_lcp}, most researchers have pivoted towards using deep learning tools for spoof detection \cite{nogueira_vgg,chugh_spoofbuster,chugh_generalization,Hi_CNN,B_2020_WACV,fld_liveness}. One popular CNN-based spoof detection scheme was Fingerprint SpoofBuster (FSB), proposed by Chugh et al. \cite{chugh_spoofbuster}. Here, the authors trained a two-class Convolutional Neural Network (CNN) using local patches around the extracted minutiae. This method proved to be highly effective for liveness detection compared to using whole fingerprint images or randomly selected local patches to ascertain liveness. Here, the system detects minutiae points and extracts patches of $96\times96$ dimensions around minutiae. The patches are then aligned and cropped before being passed into a CNN after which the patch scores are fused to generate a spoofness score. This work exceeded the performance of the SOTA at the time, with an Average Classification Error (ACER) of 0.64\% compared to 1.90\% ACER. While SpoofBuster yields highly accurate results during inference, there is an added latency involved in extracting the minutiae patches as well as an additional inference of the network per minutiae patch extracted.

\subsection{Joint Spoof Detection and Verification}
Combining spoof detection with verification has recently become a priority in secure biometric systems. One key area of focus is on trying to generate a proper metric for the fusion of the individual matching and liveness scores \cite{livdet_2019,livdet_2021}. This combined metric score should be reflective of the system's ability to accept genuine bona fide (i.e., live) fingerprint pairs (two bona fide fingerprint images belonging to the same finger) and reject imposter fingerprint pairs, of which there are two types:
\begin{enumerate}
    \item Zero-effort imposter attacks: two bona fide fingerprint images belonging to two different fingers.
    \item Imposter presentation attacks: one or more of the fingerprint images being compared is from a presentation attack, which may be either of the same finger or of two different fingers.
\end{enumerate}

There have been several Joint Spoof Detection and Verification systems proposed for modalities such as Face \cite{Chingovska_joint_operation, al2023unified}, Voice \cite{sizov_speaker} and Iris \cite{dhar_eyepad}. In the realm of fingerprint recognition, the Fingerprint Liveness Detection competitions since 2019 have focused on developing integrated fingerprint spoof detection and matching systems that utilize score-level fusion as means of outputting a combined metric score. The solutions presented at these competitions revolved around a parallel deployment of fingerprint anti-spoofing and verification, where both networks need not share the same base network and could operate independently of each other. 

Recently, Popli et al. \cite{popli_unified} proposed a unified spoof detection and matching model which accomplished both fingerprint spoof detection and matching with accuracy comparable to prior stand-alone methods. Popli et al. also showed that this system reduces the memory and time latency by $50\%$ and $40\%$ compared to existing systems. However, this work did not address the accuracy of the network in generating a combined metric for both spoof detection and matching and also employed minutiae-based spoof detection and fingerprint matching approaches, which are typically much slower than their deep learning counterparts. Furthermore, when evaluating the performance of the model put forward by Popli et al., it can be observed that they did not perform the evaluations on liveness detection and matching accuracy upon the same dataset, and thus we are not able to discern how well the system can distinguish genuine matches from a combination of zero-effort impostor attacks and presentation attacks. Therefore, it is impossible to evaluate the accuracy of detecting genuine matches by this system as liveness and matching accuracy were only evaluated independently. Hence, there arises a need for a system that is computationally efficient and maintains high accuracy. A summary of the different fingerprint joint spoof detection and matching algorithms can be observed in Table \ref{table_summary}.

\begin{figure*}[!t]
\includegraphics[width=\linewidth]{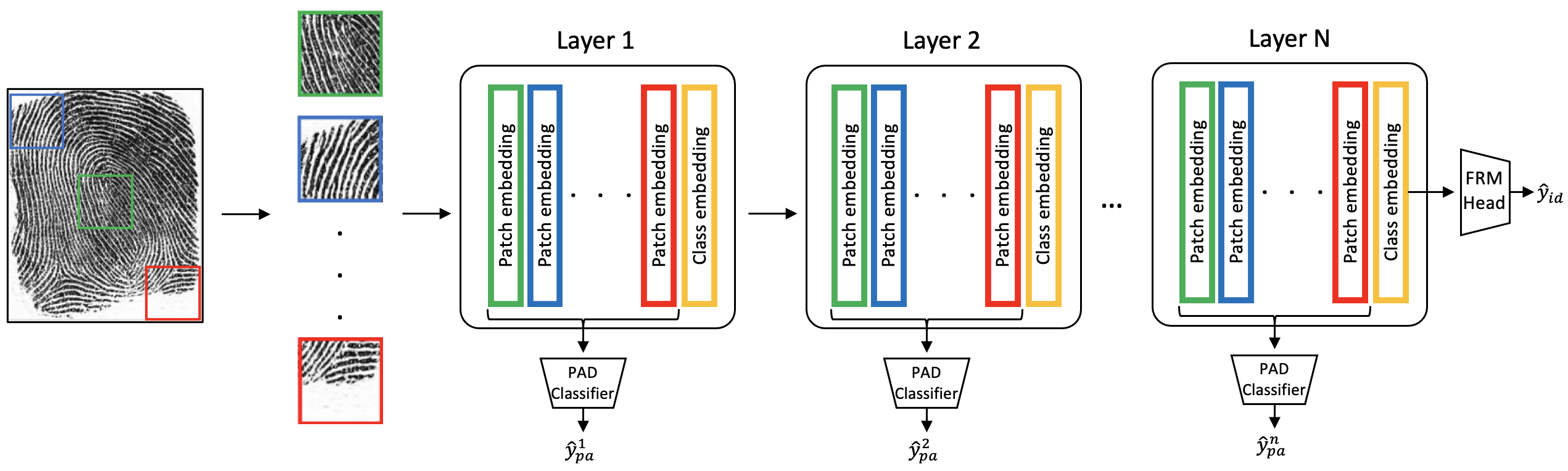} 
\caption{Overview of the ViT Unified model architecture.}
\label{fig:architecture}
\end{figure*}

\section{Methodology}
Our approach to unified fingerprint recognition and presentation attack detection (PAD) begins with a sequential system utilizing two ViT models, one tasked with recognition and the other for PAD. Next, we combine the two architectures into a single ViT model, thereby significantly reducing the number of parameters and latency of the end-to-end system. The following sections will detail each component of our networks.

\subsection{ViT Recognition Model}
A vanilla ViT backbone consists of N transformer blocks, with each block containing multi-head self attention (MHSA), multi-layer perceptron (MLP) layers, and residual connections~\cite{dosovitskiy2020image}. An input image is first split into non-overlapping patches, where each patch is embedded into a fixed-length representation and assigned a positional embedding to account for its location in the original image. In addition to the patch embeddings, a classification embedding is usually concatenated to the input to serve as a global representation of the entire feature set. This global representation has been shown to be incredibly useful in predicting fingerprint identity in several previous papers on fingerprint recognition utilizing ViT models~\cite{tandon2022transformer, grosz2022minutiae, grosz_afrnet}. Thus, we employ the ViT architecture as our fingerprint recognition model (FRM). In the rest of the paper we refer to this model as ViT FRM.

For our specific implementation, we utilize the small ViT version with input image dimensions of 3$\times$224$\times$224, patch size of 16, number of attention heads of 6, and layer depth of 12. We selected this architecture as it presents an adequate trade-off in speed and accuracy compared to other ViT variants. The model is trained for recognition via an ArcFace loss function with a margin of 0.5, learning rate of 1e-4, weight decay of 2e-5, polynomial learning rate decay function with a power of 3 and minimum learning rate of 1e-5, and batch size of 256 across four Nvidia GeForce RTX 2080 Ti GPUs. We utilized the AdamW optimizer~\cite{loshchilov2017decoupled} and trained the model for 58 epochs. The model was initialized using the pre-trained ImageNet weights made available by the open-sourced pytorch-image-models git repository~\cite{rw2019timm}. This model was trained on a combination of MSP~\cite{yoon2015longitudinal}, NIST SD 302~\cite{sd302}, PrintsGAN~\cite{engelsma2022printsgan}, SpoofGAN~\cite{grosz2022spoofgan}, and MSU Self-Collection fingerprint datasets.

\subsection{ViT PAD Model}
Previous studies on presentation attack detection for both fingerprints~\cite{chugh_spoofbuster} and face~\cite{deb2020look, al2023unified} modalities suggest that local features are more well suited for PAD. Therefore, in this work, we propose to utilize the local patch embeddings encoded by ViT for this task, which we refer to as ViT PAD. In particular, we append a PAD classifier consisting of two MLP layers during training to the output of each intermediate transformer block to regress to a classification embedding of dimension 2 to classify between input bona fide images and presentation attacks. Since adding a PAD classifier to each block in the architecture greatly increases the number of parameters of the model, we conduct additional experiments to decide which layer(s) and how many layers do we need to use in the classification to achieve the best trade-off in computational resources and PAD performance. This analysis is given in section~\ref{sec:ablation}.

For our specific implementation, we again utilize the small ViT version with input image dimensions of 3$\times$224$\times$224, patch size of 16, number of attention heads of 6, and layer depth of 12. The model is trained for PAD via a Cross Entropy loss function utilizing 12 MLP classifiers (one for each block), learning rate of 1e-4, weight decay of 2e-5, polynomial learning rate decay function with a power of 3 and minimum learning rate of 1e-5, and batch size of 256 across four Nvidia GeForce RTX 2080 Ti GPUs. Again, we utilized the AdamW optimizer~\cite{loshchilov2017decoupled}, training for 19 epochs. The model was initialized using the pre-trained ViT model for recognition mentioned above and finetuned on SpoofGAN, MSU Self-Collection, LivDet 2013 CrossMatch, LivDet 2015 CrossMatch, MSU FPAD, and MSU FPADv2 training datasets. In total, there was 26,120 PA images and 89,241 bona fide (i.e., live) images in training.

\subsection{ViT Unified Architecture}
To unify the two tasks (PAD and recognition), we use a single ViT model (ViT Unified) to perform both tasks. In particular, we train a ViT model with intermediate PAD classifiers after each transformer block and perform the recognition task with a softmax layer operating on the classification embeddings produced by final layer of the network. Since there are no large-scale databases of paired live and PA images, we do not have the luxury of easily training for both tasks at one time, as the model would severely overfit on the paucity of data offered by, for example, the LivDet datasets (which usually have just a few thousand images). Instead, we use our pretrained ViT recognition model as a teacher model during training of our unified model. Thus, we use the classification embedding from our teacher models as the ground truth supervision via an MSE loss with the predicted classification embeddings by our unified model. Again, we append an MLP classifiers to the output patch embeddings of each transformer layer and utilized a cross entropy loss between the predicted PA score and the ground truth bona fide/PA labels. An overview of the unified model architecture is given in Figure\ref{fig:architecture}.

This model is trained for both tasks with a learning rate of 1e-4, weight decay of 2e-5, polynomial learning rate decay function with a power of 3 and minimum learning rate of 1e-5, and batch size of 256 across four Nvidia GeForce RTX 2080 Ti GPUs utilizing the AdamW optimizer~\cite{loshchilov2017decoupled} for 19 epochs. The model was initialized using the trained ViT model for recognition mentioned above and finetuned jointly for both tasks.

\section{Experimental Results}
In this section, we give details on all training and evaluation datasets used, define the protocols and metrics used to evaluate individual task performance, as well as combined performance, and present the results obtained for both our sequential dual-ViT system and the proposed unified architecture against several baseline methods.

\subsection{Databases}
All training and test datasets used in this study are listed in Table~\ref{tab:datasets}, along with the number of unique fingers and number of bona fide and PA images. For evaluation purposes, we are using LivDet 2013 and LivDet 2015. We have chosen to limit this study to the CrossMatch sensor of both datasets for the following reasons: (i) both LivDet datasets contain a diverse set of fingerprint readers which employ different sensing technology, thus introducing a cross-sensor evaluation component, which is outside the scope of this study and (ii) by nature, ViT models require massive amount of data in order to avoid over fitting and the only publicly available fingerprint spoof datasets of reasonable size utilize a CrossMatch sensor (MSU FPAD, MSU FPADv2, and SpoofGAN). We leave it to future work to address the challenges of limited data and cross-sensor generalization in order to adapt our architecture to additional sensors and datasets.

\begin{table}
\footnotesize
\centering
\caption{Fingerprint datasets used in this study. Train and test splits are disjoint in identity.}
\label{tab:datasets}
\begin{tabular}{lccc}
\noalign{\hrule height 1.5pt}
\textbf{Train Dataset} & \textbf{\# Fingers} & \specialcell{\textbf{\# Live}\\\textbf{Images}} & \specialcell{\textbf{\# PA}\\\textbf{Images}}\\
\noalign{\hrule height 1.0pt}
MSP$^\dagger$~\cite{yoon2015longitudinal} & 37,411 & 447,988 & 0\\
\hline
NIST SD 302~\cite{sd302} & 1,600 & 20,008 & 0\\
\hline
PrintsGAN~\cite{engelsma2022printsgan} & 34,985 & 524,775 & 0\\
\hline
SpoofGAN~\cite{grosz2022spoofgan} & 5,082 & 18,425 & 4,990\\
\hline
MSU Self-Collection$^\dagger$ & 4,582 & 57,813 & 7,745\\
\hline
\leftcell{LivDet 2013 CM}~\cite{ghiani2013livdet} & 500 & 1,250 & 1,000 \\
\hline 
\leftcell{LivDet 2015 CM}~\cite{livdet2015} & 500 & 1,510 & 1,473 \\
\hline 
\leftcell{MSU FPAD}~\cite{chugh_spoofbuster} & N/A & 4,500 & 6,000 \\
\hline 
\leftcell{MSU FPADv2}~\cite{chugh2019fingerprint} & N/A & 5,743 & 4,912 \\
\noalign{\hrule height 1.0pt}
\textbf{Test Dataset} & \textbf{\# Fingers} & \specialcell{\textbf{\# Live}\\\textbf{Images}} & \specialcell{\textbf{\# PA}\\\textbf{Images}}\\
\noalign{\hrule height 1.0pt}
\leftcell{LivDet 2013 CM}~\cite{ghiani2013livdet} & 440 & 1,250 & 1,000 \\
\leftcell{LivDet 2015 CM}~\cite{livdet2015} & 500 & 1,500 & 1,448 \\
\noalign{\hrule height 1.5pt}
\multicolumn{4}{p{0.95\linewidth}}{$^\dagger$ Not publicly available. MSP database is an operational forensic dataset which cannot be released for privacy reasons.}\\
\end{tabular}
\end{table}

\subsection{Protocols and Metrics}
We use the following ISO standard accuracy metrics~\cite{isopad}, some of which were used in the LivDet competition series; namely, in the 2019\cite{livdet_2019} and 2021\cite{livdet_2021} editions which focused on joint model performance. These metrics are given below:
\begin{enumerate}
    \item \textbf{Attack Presentation Classification Error Rate (APCER)}: Proportion of attack presentations using the same PAI species incorrectly classified as bona fide presentations. Note, for simplicity in our tables, we report average APCER across all PAI species given in each dataset.
    \item \textbf{Bona fide Presentation Classification Error Rate (BPCER)}: Proportion of bona fide presentations incorrectly classified as presentation attacks.
    \item \textbf{Average Classification Error Rate (ACER)}: The mean of APCER and BPCER.
    \item \textbf{FNMR (False Non-Match Rate)}: Rate of genuine bona fide matches classified as impostors. 
    \item \textbf{FMR (False Match Rate)}: Rate of zero-effort impostors classified as genuine. 
    \item \textbf{IAPMR (Impostor Attack Presentation Match Rate)}: Proportion of imposter attack presentations using the same PAI species in which the target reference is matched. Note, for simplicity in our tables, we report average IAPMR across all PAI species given in each dataset.
    \item \textbf{Integrated Matching (IM) Accuracy}: Percentage of samples correctly recognized by the joint system. This is taken as 1 - (FNMR+FMR+IAPMR)/3.
        
\end{enumerate}

\subsection{Joint Authentication and PAD Performance}
We compare the authentication performance of our models to two state-of-the-art fingerprint recognition algorithms, DeepPrint~\cite{engelsma_deepprint} and the commercial Verifinger v12.3 algorithm by Neurotechnology. The results given in Table~\ref{tab:match_acc} demonstrate that our trained ViT recognition model outperforms DeepPrint and gives very competitive performance with Verifinger, at a dramatic reduction in latency (4.69 ms for ViT compared to 1,066 ms for Verifinger). Furthermore, our ViT model has just 27.7\% the amount of parameters as DeepPrint (21.83 M compared to 78.81 M, respectively). Our ViT Unified model also demonstrates competitive matching performance with only a slight decrease in matching accuracy compared to the specialized ViT recognition model (99.09\% vs. 99.42\% TAR @ FAR=0.01\% on LivDet 2013 and 99.73\% vs. 99.80\% on LivDet 2015, respectively).

In terms of presentation attack detection, we compare our models with the state-of-the-art SpoofBuster model designed by Chugh et al.~\cite{chugh_spoofbuster}. For a fair comparison with our models, we have implemented and trained our own SpoofBuster models based off the Inception-v3 backbone on the same training datasets used to train our ViT PAD and ViT Unified models. According to the results shown in Table~\ref{tab:pad_acc}, our specialized ViT PAD model achieves competitive performance with SpoofBuster (1.91\% vs. 1.07\% ACER on LivDet 2013 CrossMatch and 0.51\% vs. 0.48\% ACER on LivDet 2015 CrossMatch, respectively) but at a significant reduction in latency (2.53 ms vs. 466.12 ms, respectively). The reduction in latency is mainly due to SpoofBuster's patch-based approach, where an individual inference from the network is required for each minutiae patch extracted from the input images. Using the average number of 41 minutiae extracted on the LivDet 2015 CrossMatch dataset and an inference speed of Inception-v3 network of 10.84 ms per minutiae patch, the latency for SpoofBuster becomes about 466.12 ms per fingerprint image. 

For evaluating the joint PAD and recognition performance of our unified system, we benchmark our algorithms against two joint system architectures: (i) a combination of Verifinger and SpoofBuster and (ii) a combination of DeepPrint and SpoofBuster. Shown in Table~\ref{tab:joint_acc}, our sequential system design of ViT FRM plus ViT PAD and our ViT Unified system design both achieve competitive performance with the other baselines, but again at a significant reduction in parameters and latency. In particular, our systems come within 1\% of the performance of either baseline method but are more than 100$\times$ faster in making a unified liveness and identity decision. 

\begin{table*}
\footnotesize
\centering
\caption{Authentication performance of the proposed ViT Unified model compared to baseline models.}
\label{tab:match_acc}
\begin{tabular}{c|ccc|ccc}
\noalign{\hrule height 1.5pt}
\multirow{2}{*}{\textbf{Model}} & \multicolumn{3}{c|}{\textbf{LivDet 2013 CM}} & \multicolumn{3}{c}{\textbf{LivDet 2015 CM}}\\ 
\cline{2-7}
& \specialcell{\textbf{TAR (\%) @}\\\textbf{FAR=0.01\%}} & \specialcell{\textbf{FMR (\%)}} & \specialcell{\textbf{FNMR (\%)}} & \specialcell{\textbf{TAR (\%) @}\\\textbf{FAR=0.01\%}} & \specialcell{\textbf{FMR (\%)}} & \specialcell{\textbf{FNMR (\%)}}\\
\noalign{\hrule height 1.0pt}
\specialcell{Verifinger v12.3} & 99.42 & 0.26 & 0.33 & 100 & 0 & 0.07\\
\hline
\specialcell{DeepPrint~\cite{engelsma_deepprint}} & 96.96 & 0.81 & 0.83 & 99.27 & 0.24 & 0.27\\
\hline 
\specialcell{ViT FRM (proposed)} & 99.42 & 0.39 & 0.41 & 99.80 & 0.14 & 0.13\\
\hline 
\specialcell{ViT Unified (proposed)} & 99.09 & 0.55 & 0.58 & 99.73 & 0.14 & 0.13\\
\noalign{\hrule height 1.5pt}
\end{tabular}
\end{table*}

\begin{table}
\footnotesize
\centering
\caption{Presentation attack detection performance of the proposed ViT Unified model compared to baseline models. Results reported in Average Classification Error Rate (ACER).}
\label{tab:pad_acc}
\begin{tabular}{c|c|c}
\noalign{\hrule height 1.5pt}
\multirow{1}{*}{\textbf{Model}} & \multicolumn{1}{c|}{\textbf{LivDet 2013 CM}} & \multicolumn{1}{c}{\textbf{LivDet 2015 CM}}\\ 
\noalign{\hrule height 1.0pt}
\specialcell{SpoofBuster~\cite{chugh_spoofbuster}} & 1.07 & 0.51\\
\hline 
\specialcell{ViT PAD (proposed)} & 1.91 & 0.48\\
\hline 
\specialcell{ViT Unified (proposed)} & 1.91 & 0.48\\
\noalign{\hrule height 1.5pt}
\end{tabular}
\end{table}

\begin{table}
\footnotesize
\centering
\caption{Integrated Matching (IM) accuracy of our ViT Unified model compared to baseline models.}
\label{tab:joint_acc}
\begin{tabular}{ccccc}
\noalign{\hrule height 1.5pt}
\textbf{Model} & \specialcell{\textbf{\# Params.}\\\textbf{(M)}} & \specialcell{\textbf{Latency}\\\textbf{(ms)$^\ddagger$}} & \specialcell{\textbf{LivDet}\\\textbf{2013 CM}} & \specialcell{\textbf{LivDet}\\\textbf{2015 CM}}\\
\noalign{\hrule height 1.0pt}
\specialcell{Verifinger v12.3 +}\\{SpoofBuster~\cite{chugh_spoofbuster}} & \specialcell{N/A} & \specialcell{1066} & \specialcell{98.94} & \specialcell{99.93}\\
\hline
\specialcell{DeepPrint~\cite{engelsma_deepprint} +}\\{SpoofBuster~\cite{chugh_spoofbuster}} & 102.64 & 491.23 & 98.59 & 99.79\\
\hline 
\specialcell{ViT FRM + ViT}\\{PAD (proposed)} & 43.81 & 8.08 & 98.21 & 99.68\\
\hline 
\specialcell{ViT Unified}\\{(proposed)} & 21.98 & 4.69 & 98.03 & 99.70\\
\noalign{\hrule height 1.5pt}
\multicolumn{5}{p{0.95\linewidth}}{$^\ddagger$ Computed on an Nvidia RTX A6000 GPU with batch size of 1.}\\
\end{tabular}
\end{table}

\section{Discussion}
In this section we discuss some ablation analysis related to our methods, the performance vs. latency trade-off in our unified vs. sequential architectures, and some possible future directions to improve upon our work.

\subsection{Ablation Analysis}
\label{sec:ablation}
When training our ViT model for PAD, we introduced MLP classifiers at the output of each intermediate transformer layer; however, this greatly adds to the number of parameters needed to store the parameters of all the linear layers involved. Therefore, we conducted an ablation analysis to determine which layer(s) contribute most to the final spoof detection in order to reduce the overall number of parameters required by the model. This analysis is shown in Table~\ref{tab:ablation}. We find that layers 5, 6, 8, and 10 have the highest PAD performance across LivDet 2013 and LivDet 2015 CrossMatch sensors, with layer 6 having the best average performance. For this reason, during inference for our final ViT PAD model and ViT Unified models, we just load the parameters corresponding to layer 6 for computing the PAD result. As was noted in \cite{al2023unified}, we can also use an ensemble of PAD classifiers during the evaluation to improve the performance slightly (e.g., combining the outputs of layers 5, 6, 8, and 10); however, we found that the modest accuracy improvement of less than 1\% most likely does not warrant the 3$\times$ increase in model parameters.

\begin{table*}
\footnotesize
\centering
\caption{Ablation analysis on PAD classifier position within ViT intermediate layers. Results are given in APCER @ BPCER=0.2\%.}
\label{tab:ablation}
\begin{tabular}{c|c|c|c|c|c}
\noalign{\hrule height 1.5pt}
\multirow{2}{*}{Layer} & \multicolumn{2}{c|}{ViT PAD}                    & \multicolumn{2}{c|}{ViT Unified}                & \multicolumn{1}{c}{\multirow{2}{*}{Average}} \\
\cline{2-5}
                          & LivDet 2013 CrossMatch & LivDet2015 CrossMatch & LivDet 2013 CrossMatch & LivDet2015 CrossMatch & \multicolumn{1}{c}{}\\
\noalign{\hrule height 1.0pt}
1                         & 100.0                  & 100.0                 & 100.0                  & 100.0                 & 100.0                                        \\
2                         & 50.4                   & 15.1                  & 47.9                   & 23.1                  & 34.1                                         \\
3                         & 21.8                   & 3.3                   & 22.6                   & 5.9                   & 13.4                                         \\
4                         & 9.3                    & 1.5                   & 14.1                   & 1.9                   & 6.7                                          \\
5                & \textbf{5.7}           & 1.1                   & 11.3                   & 0.9                   & 4.8                                          \\
6                & 7.1                    & 0.8                   & \textbf{7.4}           & 0.8                   & \textbf{4.0}                                 \\
7                & 8.3                    & 1.4                   & 10.4                   & 0.8                   & 5.2                                          \\
8                & 8.5                    & \textbf{0.7}          & 9.7                    & 0.8                   & 4.9                                          \\
9                         & 8.5                    & 1.1                   & 12.8                   & 1.0                   & 5.8                                          \\
10                        & 9.6                    & 1.1                   & 12.9                   & \textbf{0.6}          & 6.1                                          \\
11                        & 7.9                    & 0.8                   & 13.3                   & 0.9                   & 5.7                                          \\
12                        & 7.9                    & 0.8                   & 13.3                   & 0.9                   & 5.7                                         \\
\noalign{\hrule height 1.5pt}
\end{tabular}
\end{table*}

\subsection{Performance Trade-off: Unified vs. Sequential Architectures}
As alluded to in the previous experimental results section, we observed only marginal decrease in accuracy for both PAD and recognition tasks when moving from a sequential system consisting of two separate ViT models to a unified system of just one ViT model. For example, the average IM accuracy of our sequential systems was 98.95\% compared to 98.87\% for the unified system, but at roughly a cost of 2$\times$ the parameters and $2\times$ the latency. Depending on the application (e.g., deployment on an edge device like a mobile phone), this speed up in latency at very modest reduction in accuracy may be desirable.

\subsection{Future Work}
Even though our models demonstrate competitive performance with baseline systems at a significant saving in computation requirements (both in model size and latency), there are still future improvements that could be done to improve the accuracy. First, we found that ViT's require large amounts of PA training data to avoid over fitting. A potential future direction would be to try to leverage more synthetic data or design better training strategies to overcome the paucity of publicly available fingerprint spoof data. Additionally, as the authors noted in \cite{chugh_spoofbuster}, fingerprint spoof detection performance was improved via incorporation of fingerprint domain knowledge (e.g., minutiae) into the model architecture. As it stands, our vanilla application of ViT does not leverage any fingerprint domain knowledge to improve the performance, both for PAD and for recognition. 

\section{Conclusion}
In this work, we presented two solutions to joint fingerprint recognition and presentation attack detection; namely, a sequential system involving two separate ViT models and a ViT Unified model that performs both tasks concurrently. We demonstrated competitive performance of both models at a significant reduction in parameters and latency for both recognition and presentation attack detection compared to state-of-the-art baseline models (Verifinger v12.3 + SpoofBuster and DeepPrint + SpoofBuster). As part of future work, our model can be extended to additional datasets and sensors and may incorporate fingerprint domain knowledge (e.g., minutiae) to improve both the PAD and recognition performance.

{\small
\bibliographystyle{ieee}
\bibliography{egbib}
}

\end{document}